\documentclass{bmvc2k}

\title{Dual-Stream Attention with Multi-Modal Queries for Object Detection in Transportation Applications}

\addauthor{Noreen Anwar}{noreen.anwar@polymtl.ca}{1}
\addauthor{Guillaume-Alexandre Bilodeau}{gabilodeau@polymtl.ca}{1}
\addauthor{Wassim Bouachir}{wassim.bouachir@teluq.ca}{2}

\addinstitution{
LITIV, Polytechnique Montréal\\
Montréal, Canada
}

\addinstitution{
Data Science Laboratory,\\
Université du Québec (TELUQ)\\
Montréal, Canada
}

\runninghead{Student, Prof, Collaborator}{BMVC Author Guidelines}

\begin{document}

\maketitle

\begin{abstract}
Transformer-based object detectors often struggle with occlusions, fine-grained localization, and computational inefficiency caused by fixed queries and dense attention. We propose DAMM, Dual-stream Attention with Multi-Modal queries, a novel framework introducing both query adaptation and structured cross-attention for improved accuracy and efficiency. DAMM capitalizes on three types of queries: appearance-based queries from vision-language models, positional queries using polygonal embeddings, and random learned queries for general scene coverage. Furthermore, a dual-stream cross-attention module separately refines semantic and spatial features, boosting localization precision in cluttered scenes. We evaluated DAMM on four challenging benchmarks, and it achieved state-of-the-art performance in average precision (AP) and recall, demonstrating the effectiveness of multi-modal query adaptation and dual-stream attention. Source code is at: \href{https://github.com/DET-LIP/DAMM}{GitHub}.
\end{abstract}

\section{Introduction}

Traditional approaches for object detection in transportation applications focused on classifying and localizing objects within predefined categories using rigid bounding boxes. While these methods have achieved remarkable success, they struggle to generalize to real-world scenarios characterized by arbitrary objects, occlusions, and complex spatial structures. Recent transformer-based detectors, such as DETR and its variants, have improved detection accuracy, but remain constrained by static object queries, computationally expensive global attention mechanisms, and limited spatial granularity. These limitations are particularly pronounced in dynamic environments where object shapes, occlusions, and distributions deviate significantly from what was seen during training, as exemplified in aerial and urban scenarios. UAVDT~\cite{du2018Unmanned} and VisDrone~\cite{zhu2021detection} highlight the challenges of detecting small and highly occluded objects in UAV-based imagery. Existing methods, such as RT-DETR~\cite{zhang2025uav} and UAV-DETR~\cite{zhang2025uav}, attempt to optimize transformers for UAV-based detection but remain limited by their reliance on predefined object distributions.


Recent breakthroughs in vision-language models (VLMs)~\cite{liu2023grounding, du2022learning, minderer2022simple} have opened new avenues for open-world recognition by narrowing the gap between visual and textual representations. However, existing detection frameworks fail to fully capitalize on these advancements due to three shortcomings: (1) reliance on static query embeddings, which lack adaptability to diverse object appearances and contexts; (2) computational inefficiencies stemming from dense self-attention mechanisms in the decoder; and (3) the use of rigid, axis-aligned bounding box positional queries, which provide suboptimal positioning for objects with irregular geometries.  We address these shortcomings with four contributions:

\begin{itemize}
\item Multi-Modal Queries: We introduce a unified query set that integrates appearance-based queries derived from vision-language embeddings to capture semantics, position-based queries from segmentation-driven polygonal embeddings to capture spatial information, and random learned queries to ensure robust general scene coverage. This multi-modal approach enables the model to adapt to diverse object appearances and contexts dynamically, addressing the first shortcoming.

\item Adaptive Query Fusion: We designed a learnable mechanism to dynamically refine static and dynamic queries within the transformer decoder. This adaptive fusion accelerates convergence while improving generalization across diverse detection scenarios, complementing our first contribution.

\item Dual-Stream Cross-Attention: We propose a structured attention mechanism that decouples semantic and spatial representations, optimizing feature alignment from unified query adaptation while significantly reducing computational overhead. This dual-stream design enhances the models ability to reason about both object semantic and spatial relationships efficiently, addressing the second shortcoming.

\item Polygonal Positional Embeddings: We introduce an original scheme that encodes object boundaries as polygonal positional embeddings, enabling our model to precisely capture irregular shapes and occlusions, thereby achieving superior localization accuracy, and addressing the third shortcoming.
\end{itemize}

These contributions are integrated in DAMM (Dual-Stream Attention with Multi-Modal Queries), a novel transformer-based video detection framework fusing appearance queries from Grounding DINO \cite{li2021grounded}, positional queries from SAM \cite{kirillov2023segment}, and random learned queries into a unified representation. These queries are processed via a dual-stream decoder that separately refines semantic and spatial cues to enhance detection accuracy across diverse scenarios. We evaluated DAMM on four challenging datasets, Cityscapes~\cite{Cordts2016Cityscapes}, UAVDT~\cite{du2018Unmanned}, VisDrone~\cite{zhu2018visdrone}, and UA-DETRAC~\cite{Wen2020UA-DETRAC}, demonstrating state-of-the-art performance in both mean average precision (mAP) and recall. DAMM consistently outperforms existing methods, particularly in scenarios involving occlusions and fine-grained spatial structures. 




\section{Related Work}

Object detection has evolved significantly with the advent of transformer and vision-language models. Despite these advancements, existing methods still face challenges in query representation, spatial precision, and cross-modal feature integration. Our work builds upon recent developments in structured detection models, multi-modal fusion, and query refinement strategies to introduce a more flexible and robust detection framework.

\noindent\textbf{Transformer-Based Object Detection:}
The introduction of DETR~\cite{carion2020end} transformed object detection by eliminating traditional heuristics, such as non-maximum suppression (NMS) and anchor generation. However, its slow convergence and global dense attention motivated the development of more efficient variants. Deformable DETR~\cite{zhu2020deformable} significantly improved computational efficiency by introducing sparse spatial sampling, while DINO~\cite{zhang2022dino} and DN-DETR~\cite{li2022dn} refined query-based learning (that is, how object queries are generated, updated, and refined during training) to accelerate training and enhance feature refinement. Other methods, such as Conditional DETR~\cite{meng2021conditional}, introduced conditional queries to better align detection predictions, while Sparse DETR~\cite{zhang2025uav} and QueryDet~\cite{yang2022querydet} focused on optimizing similar queries efficiency by propagating only the most informative ones. Despite these advances, existing DETR-based models remain limited by static object queries that lack adaptability in dynamic scenes. Our approach addresses this issue by incorporating multi-modal query adaptation, allowing queries to dynamically fuse semantic and spatial embeddings for better detection performance.

\noindent\textbf{Spatial Representation Learning:}
Traditional bounding boxes often struggle with irregular shapes in aerial (VisDrone \cite{zhu2018visdrone}) and urban (Cityscapes \cite{Cordts2016Cityscapes}) datasets. Deformable DETR~\cite{zhu2020deformable} enhanced spatial sensitivity through offset learning, while SDPDet~\cite{yin2024sdpdet} improved with scale-aware localization by dynamically fusing multi-scale features and incorporating scale-specific modulation into its regression head, enabling more accurate detection of objects with varying sizes. Recent contour-based methods like Poly-YOLO~\cite{sochor2022poly} and PolarMask~\cite{xie2020polarmask} demonstrated the benefits of polygonal representations, but require complex post-processing. We address this by integrating segmentation-derived polygonal embeddings directly into the transformer decoder, achieving more precise localization without additional refinement steps.

\noindent\textbf{Vision-Language Integration in Detection:}
Recent vision-language models (VLMs) have extended detection capabilities beyond closed-set categories, allowing for more flexible and open-world recognition. CLIP~\cite{radford2021learning} and ALIGN~\cite{ji2024align} demonstrated strong zero-shot classification capabilities, leading to their adaptation in detection frameworks such as OWL-ViT~\cite{minderer2022simple} and GLIP~\cite{li2022grounded}. Grounding DINO~\cite{liu2023grounding} further incorporated text-driven region supervision to enhance detection performance. However, most existing methods rely on static text embeddings, limiting their adaptability to unseen categories. OV-DETR~\cite{zang2022open} addressed this by integrating open-vocabulary learning into a DETR-based framework. However, it still lacks dynamic query adaptation. Our model builds on these works by introducing adaptive query fusion, which integrates both vision-language embeddings and structured spatial cues, leading to improved generalization across diverse object categories.

\section{Methodology}

\begin{figure*}[t]
    \centering
    \includegraphics[width=0.7\linewidth]{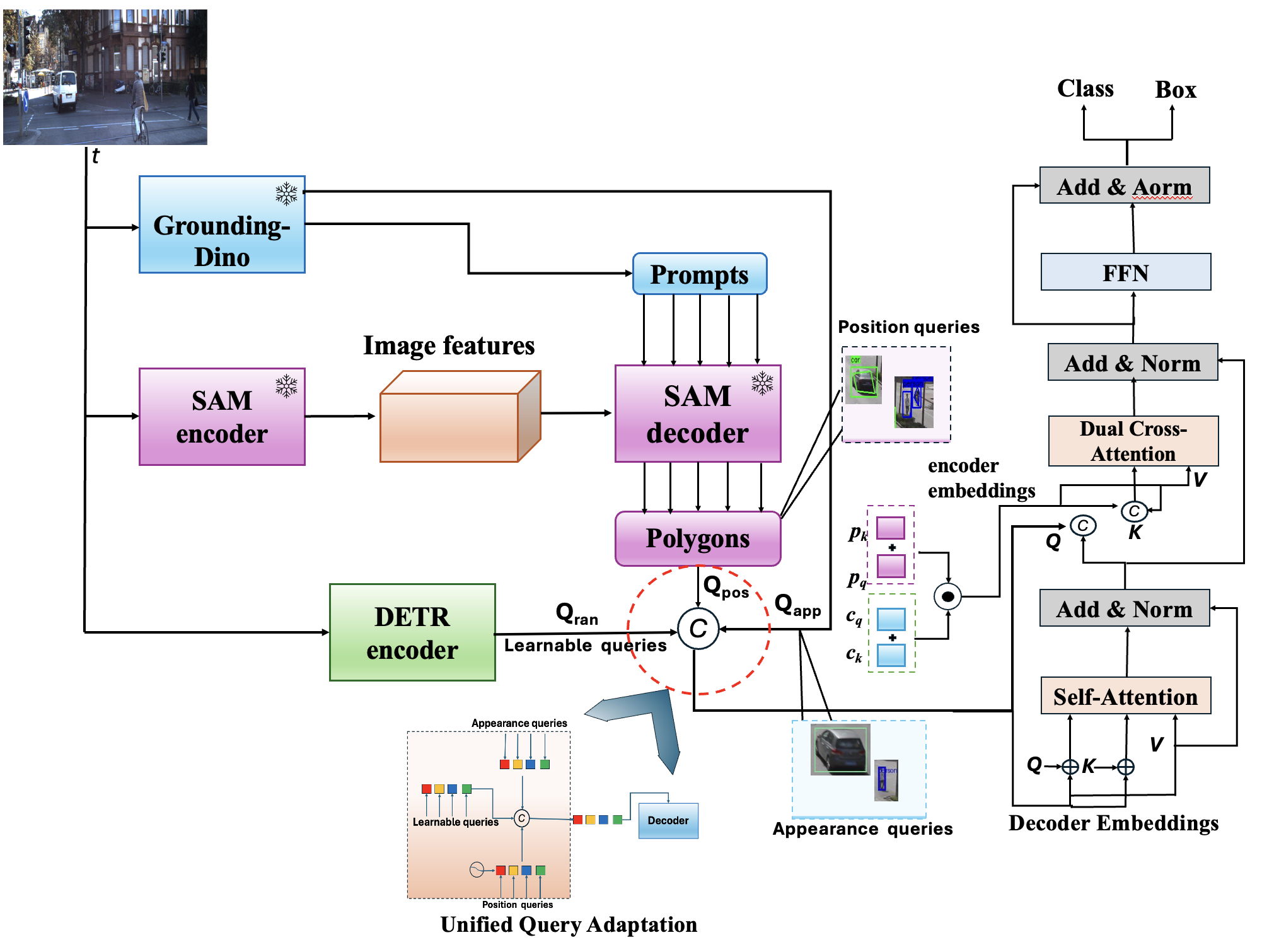}
    \caption{\textbf{DAMM Framework.} Our approach builds upon transformer-based detection by integrating multi-modal queries, unified query adaptation fusion, and dual-stream cross-attention. Object queries dynamically incorporate appearance-based and positional cues.}
    \label{fig:model_architecture}
\end{figure*}

To improve generalization, DAMM introduces adaptive query mechanisms that dynamically refine representations across multiple detection stages. By combining multi-modal feature fusion, fine-grained embeddings, and structured attention mechanisms, DAMM achieves superior robustness across diverse benchmarks, as demonstrated in our results. DAMM overcomes the constraints of fixed object queries by dynamically integrating multi-modal cues and decoupling semantic and spatial processing through a dual-stream cross-attention mechanism. As illustrated in Figure~\ref{fig:model_architecture}, DAMM fuses image semantic features from Grounding DINO, positional cues from SAM-generated polygons, and randomly learned embeddings to generate dual-stream queries that are appearance-based and position-based. By fusing these representations within a unified query adaptation mechanism, we enable the DETR-like decoder to generate more accurate object predictions. These queries undergo independent refinement via a dual cross-attention module. This design enhances detection robustness by dynamically adapting queries and leveraging multi-modal information for improved localization and recognition.

\setlength{\parskip}{0pt}

\subsection{Multi-Modal Queries}

Conventional DETR-based detectors rely on fixed object queries, limiting their adaptability to variations in object appearance and spatial configuration. To address this, we propose unified query adaptation, which constructs a dynamic query representation by fusing cues from multiple modalities. Our formulation consists of three distinct query types: 

 \noindent1) \textbf{Appearance-Based Queries} (\(\mathbf{Q}_{\text{app}}\)): Leveraging Grounding DINO \cite{li2021grounded} strengths in open-vocabulary recognition, we compute the appearance-based queries for a given image \(\mathbf{I}\) and text prompt \(\mathcal{T}\) as:
    \begin{equation}
        \mathbf{Q}_{\text{app}} = \phi_{\text{proj}}(\text{GDINO}(\mathbf{I}, \mathcal{T}));
    \end{equation}

   \noindent2) \textbf{Positional Queries} (\(\mathbf{Q}_{\text{pos}}\)): These are obtained from segmentation masks generated by SAM~\cite{kirillov2023segment} based on prompts, $\mathcal{P}$, from Grounding DINO. A polygonal approximation \(\mathbf{B}\) of each mask \(\mathbf{M}\) is made and transformed into an embedding. The polygonal representation efficiently captures the boundary information of a potential object to detect, making those queries more apt to deal with occlusions. By converting \(\mathbf{B}\) into a flattened form and processing it through an MLP, we encode rich geometric details that enhance the models ability to localize objects. For a polygonal approximation \(\mathbf{B}\), we have:
  \begin{equation}
    \mathbf{Q}_{\mathrm{pos}}
    = \mathrm{MLP}\bigl(\mathrm{Flatten}(\mathbf{B})\bigr),
\end{equation}
\noindent where \(\mathbf{M} = \mathrm{SAM}(\mathbf{I}, \mathcal{P})\), \(\mathbf{B} = \mathrm{PA}(\mathbf{M})\), and \(\mathrm{PA}(\cdot)\) computes the polygonal approximation of the mask.

     \noindent 3) \textbf{Random Learnable Queries} (\(\mathbf{Q}_{\text{ran}}\)): These are learnable queries inherited from the DETR framework, initialized from a Gaussian distribution:
    \begin{equation}
        \mathbf{Q}_{\text{ran}} \sim \mathcal{N}(0, \mathbf{I})
    \end{equation}
    Although DETR treats these queries as learnable, their initialization from a Gaussian distribution ensures diverse starting points, which are refined during training.

The overall query set is formed by concatenating the three components:
\begin{equation}
    \mathbf{Q}_{\text{combined}} = \left[ \mathbf{Q}_{\text{app}} \,\Vert\, \mathbf{Q}_{\text{pos}} \,\Vert\, \mathbf{Q}_{\text{ran}} \right],
\end{equation}
where \(\Vert\) denotes channel-wise concatenation. This multi-scale, multi-modal representation captures both semantic and spatial information effectively.

\subsection{Transformer-Based Detection Pipeline}

DAMM adheres to a standard transformer-based detection architecture comprising an encoder, a decoder, and prediction heads for classification and localization. They can be summarized as follows.  \textbf{Encoder:}  
Feature maps are extracted from a backbone network (e.g., ResNet-50 or Swin Transformer). The encoder utilizes multi-head self-attention to capture long-range dependencies and employs deformable attention~\cite{zhu2020deformable} to efficiently sample salient spatial regions. \noindent \textbf{Decoder:}  
The decoder refines object queries through two mechanisms: 1) \textbf{Self-Attention:} Enables interaction among queries, promoting context-aware refinement and reducing redundancy and 2) \textbf{Dual-Stream Cross-Attention:} Decouples semantic and spatial processing to effectively integrate category-level and localization cues.

\subsubsection{Dual-Stream Cross-Attention}
Standard cross-attention in DETR aggregates all feature components uniformly, potentially obscuring the distinction between semantic and spatial cues. Our dual-stream cross-attention module addresses this issue by decomposing the attention mechanism into two independent streams: one that processes appearance-based queries from Grounding DINO, and another that leverages position-based queries derived from SAM polygonal embeddings. Each object query is represented as 
    $\mathbf{Q}_i = \left[\mathbf{Q}_{\text{app}}^i \; ; \; \mathbf{Q}_{\text{pos}}^i\right]$,
\noindent where \(\mathbf{Q}_{\text{app}}^i\) and \(\mathbf{Q}_{\text{pos}}^i\) denote the semantic (appearance) and spatial (positional) parts of the \(i\)-th query. These components are processed independently in their respective attention streams. The cross-attention operation for the \(i\)-th query is computed as:
\begin{equation}
    \text{Attention}(\mathbf{Q}_i, \mathbf{K}, \mathbf{V}) = \mathbf{c}_q^\top \mathbf{c}_k + \mathbf{p}_q^\top \mathbf{p}_k,
\end{equation}

\noindent where $\mathbf{c}_q = \mathbf{Q}_{\text{app}}^k$ and $\mathbf{c}_k = \mathbf{K}_{\text{app}}$ represent the semantic query and key vectors, respectively, and $\mathbf{p}_q = \mathbf{Q}_{\text{pos}}^j$ and $\mathbf{p}_k = \mathbf{K}_{\text{pos}}$ denote the spatial query and key vectors.
This decoupled formulation allows the network to concentrate on content and spatial localization independently, yielding improved performance in occluded and cluttered scenes.


\subsubsection{Unified Query Adaptation Fusion}
\label{sec:adaptive_query}

Static query embeddings limit the adaptability of transformer-based detectors in dynamic scenes. We propose \textbf{adaptive query fusion}, a mechanism that iteratively refines object queries across decoder layers. At each decoder layer $t$, the query representation $\mathbf{Q}_t$ is updated using features derived from the current cross-attended output $\mathbf{F}_t$:

\begin{equation}
    \mathbf{Q}_{t+1} = \text{FFN}\left(\text{LN}(\mathbf{Q}_t + \mathbf{F}_t)\right),
\end{equation}

\noindent where $\mathbf{F}_t$ is the cross-attention output at step $t$, $\text{LN}(\cdot)$ denotes LayerNorm, and $\text{FFN}(\cdot)$ is a two-layer feedforward network with ReLU activation. This formulation enables the queries to incorporate up-to-date contextual cues from the encoder features at each stage. Importantly, the subscript $t$ indexes decoder layers (not time steps). By continuously merging semantic and spatial context through learned fusion, DAMM dynamically adapts to scene structure and object interactions, improving robustness to occlusions and complex layouts.

\subsubsection{Reference Points}
\label{sec:refpoints}
Reference points serve as the spatial anchors for box prediction in transformer-based detection frameworks. Specifically, the decoder embedding $\mathbf{f}$ encodes the displacements of the four corners of a bounding box relative to a reference point $\mathbf{s}$. The final box is obtained by predicting these displacements in the unnormalized space and then normalizing the result to the range $[0,1]$. In the original DETR framework, the reference point is statically initialized as $\mathbf{s} = [0, 0]^\top$ for all decoder queries. In DAMM, we extend this approach by exploring two dynamic formulations for initializing reference points: 1) \textbf{Global Learnable}: Unnormalized 2D coordinates $\mathbf{s}^* \in \mathbb{R}^2$ are treated as trainable parameters, and 2) \textbf{Polygon-Predict}: Coordinates are dynamically predicted from the positional query component $\mathbf{Q}_{\text{pos}}^i$ using a feedforward network (FFN): 
    
    \begin{equation}
        \mathbf{s}_i = \text{FFN}(\mathbf{Q}_{\text{pos}}^i) = \mathbf{W}_2\left(\text{ReLU}(\mathbf{W}_1\mathbf{Q}_{\text{pos}}^i)\right),
    \end{equation}
    where the FFN consists of a learnable linear projection, ReLU activation, and a final linear layer. For cross-attention, the unnormalized coordinates $\mathbf{s}_i$ are normalized to $[0,1]^\top$ using the sigmoid function,
    $\hat{\mathbf{s}}_i = \sigma(\mathbf{s}_i)$,
ensuring alignment with the spatial dimensions of the input feature maps. Unlike Deformable DETR~\cite{zhu2020deformable}, which predicts relative offsets from learnable queries, DAMM predicts \textit{absolute} coordinates directly from SAM-derived positional embeddings $\mathbf{Q}_{\text{pos}}^j$, leveraging geometric priors for more precise initialization.

\subsection{Loss Function and Inference}
  
DAMM is trained using a set-based loss with Hungarian matching \cite{rezatofighi2019generalized}:
\begin{equation}
    \mathcal{L} = \lambda_{\text{cls}}\,\mathcal{L}_{\text{cls}} + \lambda_{\text{bbox}}\,\mathcal{L}_{\text{bbox}} + \lambda_{\text{giou}}\,\mathcal{L}_{\text{giou}},
\end{equation}
where \(\mathcal{L}_{\text{cls}}\) is the focal loss for classification, \(\mathcal{L}_{\text{bbox}}\) is the L1 loss for bounding box regression, and \(\mathcal{L}_{\text{giou}}\) is the generalized IoU loss.

During inference, DAMM generates object queries via unified query adaptation. Final detections, leveraging polygonal embeddings for enhanced spatial precision, are produced in a single forward pass, enabling real-time performance while maintaining high detection accuracy in complex environments such as urban, aerial, and occluded scenes.

\section{Experiments}
We evaluated our proposed DAMM method against state-of-the-art detection frameworks on four challenging datasets featuring road users and covering diverse scenarios and viewpoints: Cityscapes~\cite{Cordts2016Cityscapes}, UAVDT~\cite{du2018Unmanned}, VisDrone~\cite{zhu2018visdrone}, and UA-DETRAC~\cite{Wen2020UA-DETRAC}. 

\begin{table}[t]
  \centering
  \fbox{%
    \resizebox{0.95\textwidth}{!}{%
      {\small
      \begin{tabular}{c|l|cccccc}
        \toprule
        \textbf{Group} & \textbf{Method} & $AP$ & $AP_{50}$ & $AP_{75}$ & $AP_S$ & $AP_M$ & $AP_L$ \\
        \midrule

        \multirow{8}{*}{\textbf{Baseline}} 

        & DETR\textsuperscript{*}~\cite{zhang2022rtdetr} & 14.7 & 26.5 & 11.3 & 4.6 & 11.2 & 28.6 \\
        & Conditional DETR~\cite{wang2021fp} & 21.1 & 42.7 & 18.8 & 3.6 & 19.8 & 41.1 \\
        & Deformable DETR\textsuperscript{*}~\cite{zhu2020deformable} & 22.5 & 41.7 & 22.5 & 7.9 & 20.7 & 42.1 \\
        & DN-DETR\textsuperscript{*}~\cite{li2022dn} & 26.0 & 46.0 & 23.0 & 7.0 & 25.0 & 43.0 \\
        & DAB-Deformable DETR\textsuperscript{*}~\cite{liu2022dabdetr} & 27.3 & 48.0 & 24.5 & 8.2 & 26.1 & 45.3 \\
        & OV-DETR~\cite{zang2022open} & 31.5 & 54.3 & 36.2 & 11.1 & 34.5 & 56.1 \\
        & DINO-Deformable DETR\textsuperscript{*}~\cite{zhang2022dino} & 33.0 & 56.0 & 38.0 & 11.5 & 35.0 & 57.0 \\
        & Grounding DINO~\cite{liu2023grounding} & \uline{35.5} & \uline{58.7} & 39.6 & \uline{13.9} & \uline{37.5} & \uline{59.4} \\
        \midrule

        \multirow{7}{*}{\textbf{Other SOTA}} 
        & YOLOS-S~\cite{wang2021fp} & 9.8 & 25.3 & 6.1 & 1.9 & 8.1 & 20.7 \\
        & UP-DETR~\cite{wang2021fp} & 23.8 & 45.7 & 20.8 & 4.0 & 20.3 & 46.6 \\
        & DELA-DETR~\cite{wang2022anchor} & 25.2 & 46.8 & 22.8 & 6.5 & 23.8 & 44.3 \\
        & FP-DETR~\cite{wang2021fp} & 29.6 & 53.6 & 28.4 & 11.2 & 30.9 & 47.4 \\
        & ViLD~\cite{gu2021open} & 29.7 & 54.3 & \textbf{52.5} & -- & -- & -- \\
        & OWL-ViT\textsuperscript{*}~\cite{minderer2022simple} & 30.0 & 53.0 & 33.0 & 10.0 & 32.0 & 54.0 \\
        & DenseCL~\cite{wang2021dense} & 30.1 & 53.5 & 35.7 & 11.8 & 32.6 & 55.2 \\
        \midrule

        \textbf{Ours} & \textbf{DAMM} & \textbf{38.5} & \textbf{62.5} & \uline{47.5} & \textbf{16.1} & \textbf{41.5} & \textbf{65.7} \\
        \bottomrule
      \end{tabular}}%
    }
  }

   \vspace{0.5em}
  \caption{\textbf{Object detection results} in terms of average precision ($AP$) on the Cityscapes validation set (best in \textbf{bold}, second best \uline{underlined}). All methods use the same ResNet-50 backbone.{*} Methods fine-tuned by us.}
  \label{tab:cityscapes}
  
\end{table}

\begin{table}[t]
  \centering
  \fbox{%
    \resizebox{0.95\textwidth}{!}{%
      {\small
      \begin{tabular}{c|l|cccccc}
        \toprule
        \textbf{Group} & \textbf{Method} & $AP$ & $AP_{50}$ & $AP_{75}$ & $AP_S$ & $AP_M$ & $AP_L$ \\
        \midrule

        \multirow{8}{*}{\textbf{Baseline}} 

        & DN-DETR\textsuperscript{*}~\cite{li2022dn} & 26.0 & 42.0 & 26.0 & 17.0 & 26.0 & 38.0 \\
        & Deformable DETR\textsuperscript{*}~\cite{zhu2020deformable} & 27.2 & 44.1 & 28.5 & 13.5 & 25.2 & 37.8 \\
        & Sparse DETR~\cite{zhang2025uav} & 27.3 & 42.2 & -- & -- & -- & -- \\
        & DAB-Deformable DETR\textsuperscript{*}~\cite{liu2022dabdetr} & 28.0 & 45.0 & 29.0 & 18.0 & 27.0 & 39.0 \\
        & Conditional-DETR\textsuperscript{*}~\cite{meng2021conditional} & 28.5 & 43.0 & 27.8 & 20.0 & 28.0 & 41.0 \\
        & DETR\textsuperscript{*}~\cite{zhang2022rtdetr} & 29.0 & 41.2 & 27.5 & 22.1 & 27.2 & 42.9 \\
        & DINO-Deformable DETR\textsuperscript{*}~\cite{zhang2022dino} & 31.1 & 49.5 & 28.2 & 18.0 & 29.4 & 47.4 \\
        & OV-DETR~\cite{zang2022open} & 32.0 & 50.0 & 30.0 & 19.0 & 31.0 & 46.0 \\
        & \underline{Grounding DINO~\cite{liu2023grounding}} & \underline{34.9} & \underline{59.3} & \underline{40.1} & 25.0 & 39.5 & \underline{50.8} \\
        \midrule

        \multirow{7}{*}{\textbf{Other SOTA}} 
        & ClusDet~\cite{yang2019clustered} & 26.7 & 50.6 & 24.7 & 18.0 & 25.0 & 38.0 \\
        & UAV-DETR-R5~\cite{zhang2025uav} & 31.5 & 51.1 & -- & -- & -- & -- \\
        & QueryDet~\cite{yang2022querydet} & 28.3 & 48.1 & 28.8 & 19.8 & 35.9 & 40.3 \\
        & RT-DETR-R50~\cite{zhang2025uav} & 28.4 & 47.0 & -- & -- & -- & -- \\
        & OWL-ViT\textsuperscript{*}~\cite{minderer2022simple} & 29.5 & 44.5 & 30.0 & 21.0 & 29.0 & 41.0 \\
        & CZ Det~\cite{meethal2023cascaded} & 33.2 & 58.3 & 33.2 & 26.1 & \uline{42.6} & 43.4 \\
        & SDPDet~\cite{yin2024sdpdet} & 33.7 & 56.6 & 34.3 & \uline{26.7} & \textbf{42.9} & 45.7 \\
        \midrule

        \textbf{Ours} & \textbf{DAMM} & \textbf{39.5} & \textbf{63.1} & \textbf{42.3} & \textbf{26.8} & 42.2 & \textbf{52.4} \\
        \bottomrule
      \end{tabular}}%
    }
  }

   \vspace{0.5em}
  \caption{\textbf{Object detection results} in terms of average precision ($AP$) on the Visdrone validation set (best in \textbf{bold}, second best \uline{underlined}). All methods use the same ResNet-50 backbone. \textsuperscript{*} Methods fine-tuned by us.}
  \label{tab:visdrone}
\end{table}%

\begin{table}[t]
  \centering
  \footnotesize
  \setlength{\tabcolsep}{2pt}     
  \renewcommand{\arraystretch}{1.05}

  \resizebox{\linewidth}{!}{%
  \begin{tabular}{@{} l  l 
                  *{6}{c}|*{6}{c} @{}}
    \toprule
      & & \multicolumn{6}{c|}{\textbf{UAVDT}} 
          & \multicolumn{6}{c}{\textbf{UA‑DETRAC}} \\
    \cmidrule(lr){3-8} \cmidrule(lr){9-14}
    \textbf{Group} 
      & \textbf{Method} 
      & $AP$ & $AP_{50}$ & $AP_{75}$ & $AP_S$ & $AP_M$ & $AP_L$ & $AP$ & $AP_{50}$ & $AP_{75}$ & $AP_S$ & $AP_M$ & $AP_L$ \\
    \midrule
    \multirow{8}{*}{\textbf{Baseline}}
      & DETR\textsuperscript{*}~\cite{zhang2022rtdetr}
        & 11.1 & 21.0 &  9.1 &  5.3 & 10.1 & 13.6  
        & 16.3 & 27.5 & 10.1 &  6.5 & 16.8 & 23.1 \\
      & Deformable DETR\textsuperscript{*}~\cite{zhu2020deformable}
        & 13.6 & 23.2 &  9.8 &  5.7 & 14.3 & 17.6  
        & 18.5 & 35.6 & 10.4 &  7.0 & 17.1 & 25.7 \\
      & OV‑DETR~\cite{zang2022open}
        & \uline{28.3} & 35.1 & 22.4 & 12.5 & 27.5 & 24.1  
        & \uline{28.8} & 37.3 & 23.9 & 13.2 & 32.5 & 28.6 \\
      & DN‑DETR\textsuperscript{*}~\cite{li2022dn}
        & 21.1 & 33.5 & 24.1 & 12.8 & 28.8 & 26.7  
        & 21.4 & 33.9 & 24.5 & 13.1 & 29.1 & 26.9 \\
      & DAB‑Deformable DETR\textsuperscript{*}~\cite{liu2022dabdetr}
        & 22.3 & 35.1 & 25.3 & 13.2 & 30.2 & 27.4  
        & 22.6 & 35.4 & 25.6 & 13.5 & 30.5 & 27.7 \\
      & Conditional DETR~\cite{meng2021conditional}
        & 23.2 & 37.3 & 26.5 & 13.6 & 29.9 & 28.1  
        & 23.5 & 37.6 & 26.8 & 13.9 & 31.3 & 28.5 \\
      & DINO‑Deformable DETR\textsuperscript{*}~\cite{zhang2022dino}
        & 24.8 & 39.9 & 26.6 & 14.8 & 32.2 & 30.4  
        & 25.3 & 40.1 & 26.9 & 15.0 & 32.5 & 30.7 \\
      & \underline{Grounding DINO}~\cite{liu2023grounding}
        & 27.0 & \uline{40.5} & \uline{27.2} & \uline{16.1} & \uline{34.3} & \uline{32.0}
        & 27.3 & \uline{41.2} & \uline{27.2} & \uline{16.3} & \uline{34.6} & \uline{32.3} \\
    \midrule
    \textbf{Ours} 
      & \textbf{DAMM}
        & \textbf{32.5} & \textbf{43.4} & \textbf{27.5} & \textbf{16.6} & \textbf{36.3} & \textbf{34.1}
        & \textbf{32.8} & \textbf{43.7} & \textbf{27.8} & \textbf{16.9} & \textbf{36.6} & \textbf{34.4} \\
    \bottomrule
\end{tabular}%
}

   \vspace{0.5em}
  \caption{\textbf{Detection performance (AP \%)} on UAVDT and UA‑DETRAC (ResNet‑50). Best in \textbf{bold}, second best \uline{underlined}. \textsuperscript{*} Fine‑tuned by us.}
  \label{tab:uavdt_ua_detrac}
\end{table}

\noindent \textbf{Implementation Details and Training.} DAMM is built upon the DETR pipeline~\cite{carion2020end} and consists of three core modules: a multi-scale backbone, a transformer encoder-decoder with six decoder layers, and shared feedforward prediction heads for classification and localization. For all experiments, we use ResNet-50 with features extracted from C3-C5 stages to ensure a fair comparison with other SOTA methods. 
We adopt the bipartite matching scheme from Deformable DETR~\cite{zhu2020deformable} and optimize the model using AdamW.
For all datasets, we maintain DETR default hyperparameters unless otherwise specified.

\subsection{Comparison with SOTA methods}
Results are reported using various object sizes (small, medium, large) and across varying overlap thresholds (e.g., $AP_{0.75}$ and $AP_{0.5}$) to comprehensively assess detection performance. Results against SOTA methods are presented in Tables~\ref{tab:cityscapes}, ~\ref{tab:visdrone}, and ~\ref{tab:uavdt_ua_detrac}. 


DAMM achieves state-of-the-art performance across urban and aerial benchmarks. On Cityscapes, it outperforms all methods in $AP$ (38.5 vs. 35.5), $AP_{50}$ (62.5 vs. 58.7), and $AP_{L}$ (65.7 vs. 59.4), validating its robustness to occlusions and scale variations. The lesser performance in $AP_{75}$ (47.5 vs. 52.5) reflects a design trade-off favoring multi-modal query fusion over exhaustive box refinement. For aerial detection (VisDrone), DAMM dominates $AP$ (39.5 vs. 34.9) and $AP_{L}$ (52.4 vs. 50.8), demonstrating superiority in most object sizes, particularly for large ones. 
Cross-domain evaluation on UAVDT and UA-DETRAC in (Table~\ref{tab:uavdt_ua_detrac}), further highlights DAMM versatility: it achieves $AP$ 32.5 (vs. 27.0) and 32.8 (vs. 27.3), respectively, demonstrating its adaptability to both aerial and ground-level viewpoints. This generalization ability is driven by the dual-stream cross-attention mechanism, which decouples semantic and spatial processing.

\subsection{Ablation Study}
\label{sec:ablation}
We conducted ablation experiments to clarify how each component of DAMM contributes to its overall performance. Specifically, we studied: (1) the effect of multi-modal query embeddings (Grounding DINO~\cite{liu2023grounding} vs.\ SAM~\cite{kirillov2023segment} vs.\ both), (2) various ways of forming reference points and polygonal embeddings, (3) the impact of dual-stream cross-attention, and (4) the benefit of adaptive query updates. Unless otherwise noted, these experiments use ResNet-50 backbones and are evaluated on the Cityscapes validation set.

\textbf{Effect of Multi‑Modal Query Embeddings} (Table~\ref{tab:query_ablation}).  
Learnable queries alone yield 34.1 $AP$; adding appearance queries (Grounding DINO) boosts $AP$ to 36.8 (+2.7), positional (SAM) to 37.2 (+3.1), and combining all three reaches 38.5 (+4.4). This demonstrates that appearance enhances category recall, positional refines localization, and fusion is the most effective.

\textbf{Reference Points and Polygonal Embeddings} (Table~\ref{tab:refpoints}).  
Fixed reference points give 35.7 $AP$; global learnable yields 38.0 (+2.3), and Polygon‑Predict achieves 38.5 (+2.8 over fixed, +0.5 over global) by encoding precise object boundaries. The gains underline the value of dynamic, shape‑aware initialization, especially for small or irregular objects.

\textbf{Dual‑Stream vs.\ Single‑Stream Cross‑Attention} (Fig.~\ref{fig:dual_vs_single}).  
Single‑stream attention scores 36.2 $AP$, whereas our dual‑stream design, separating semantic and spatial pathways, reaches 38.5 (+2.3), markedly improving detection under occlusion and across all object scales.

\textbf{Iterative Query Updates (Adaptive Fusion)} (Table~\ref{tab:query_fusion}). 
Without fusion, $AP$ is 36.0; partial fusion (appearance only) gives 37.4 (+1.4); full adaptive fusion updates all query types in each layer and achieves 38.5 (+2.5 over no fusion, +1.1 over partial), confirming that dynamic updates best accommodate scene changes and occlusions.

\begin{table}[h]
  \centering
  \footnotesize
  \begin{minipage}[t]{0.48\linewidth}
    \centering
    \resizebox{\linewidth}{!}{%
      \begin{tabular}{l|ccc}
        \toprule
        \textbf{Method} & $AP$ & $AP_{50}$ & $AP_{75}$\\
        \midrule
        Fixed \((0,0)\)        & 35.7 & 57.4 & 37.0 \\
        Global Learnable       & 38.0 & 60.7 & 45.2 \\
        Polygon‑Predict        & \textbf{38.5} & \textbf{62.5} & \textbf{47.5} \\
        \bottomrule
      \end{tabular}%
    }
    \vspace{0.5ex}
    \captionof{table}{\textbf{Reference point strategies} on Cityscapes with ResNet‑50 backbone.}
    \label{tab:refpoints}
  \end{minipage}\hfill
  \begin{minipage}[t]{0.48\linewidth}
    \centering
    \resizebox{\linewidth}{!}{%
      \begin{tabular}{l|ccc}
        \toprule
        \textbf{Fusion Type}       & $AP$   & $AP_{50}$ & $AP_{75}$ \\
        \midrule
        No Adaptive Fusion         & 36.0 & 58.1 & 42.5 \\
        Partial Fusion             & 37.4 & 60.2 & 44.1 \\
        Full Adaptive Fusion       & \textbf{38.5} & \textbf{62.5} & \textbf{47.5} \\
        \bottomrule
      \end{tabular}%
    }
    \vspace{0.5ex}
    \captionof{table}{\textbf{Impact of iterative query updates.} We report \(AP\) (\%) on Cityscapes.}
    \label{tab:query_fusion}
  \end{minipage}
\end{table}

\newcommand{\tick}{\ding{51}}        
\newcommand{\emptyCell}{\phantom{\ding{51}}}

\newsavebox{\figbox}
\sbox{\figbox}{\includegraphics[width=0.48\linewidth]{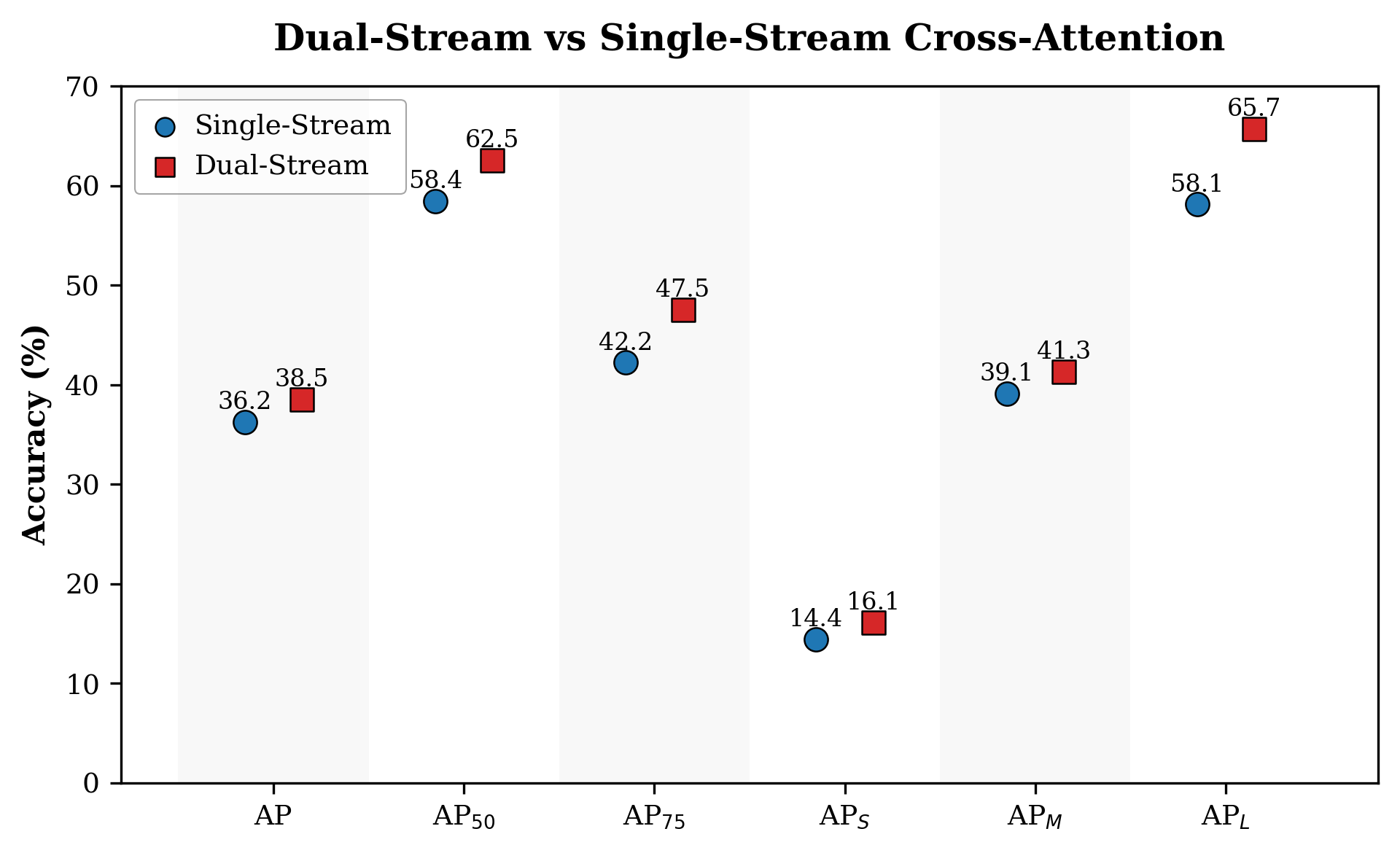}}

\begin{figure}[!htb]
  \centering
  \begin{minipage}[t][\ht\figbox][t]{0.48\linewidth}
    \vspace*{0pt} 
    \centering
    \footnotesize
    \begin{tabular}{c|c|c|c}
      \hline
      \textbf{G-DINO} & \textbf{SAM} & \textbf{RQ} & \textbf{AP (\%)} \\
      \hline
      \emptyCell      & \emptyCell   & \tick       & 34.1 \\
      \tick           & \emptyCell   & \tick       & 36.8 \\
      \emptyCell      & \tick        & \tick       & 37.2 \\
      \tick           & \tick        & \tick       & 38.5 \\
      \hline
    \end{tabular}
    \vfill        
    \captionof{table}{\footnotesize Ablation of query types: G‑DINO (appearance), SAM (positional), RQ (learnable).}
    \label{tab:query_ablation}
  \end{minipage}\hfill
  \begin{minipage}[t][\ht\figbox][t]{0.48\linewidth}
    \vspace*{-10pt} 
    \centering
    \usebox{\figbox}
    \vfill        
    \caption{\footnotesize \textbf{Single‑ vs.\ Dual‑Stream Cross‑Attention.} }
    \label{fig:dual_vs_single}
  \end{minipage}
\end{figure}

Globally, our experiments show that unified query adaptations from both Grounding DINO and SAM embeddings individually enhance performance, and combining them with random learnable queries delivers the largest gains. Moreover, dynamically predicting reference points or polygonal embeddings surpass static or globally learned baselines. Decoupling semantic and spatial attention streams consistently enhances performance, especially in crowded or occluded scenarios. Continuously refining all query embeddings through the decoding process yields better convergence and final $AP$. These findings highlight the synergy of multi-modal queries, dual-stream cross-attention, and dynamic query adaptation, all of which define DAMM advantage over conventional DETR-based detectors.

\section{Conclusion}
\label{sec:conclusion}
We introduced DAMM, a transformer-based detection framework that integrates multi-modal query adaptation and dual-stream cross-attention to enhance object detection. By dynamically fusing appearance and positional cues, DAMM improves localization and robustness in occluded and complex scenes. The overall experimental results demonstrate consistent improvements over state-of-the-art methods, highlighting the effectiveness of structured query adaptation and polygon embeddings in detection. 

\section{Acknowledgment}
We acknowledge the support of the Natural Sciences
and Engineering Research Council of Canada (NSERC),
[funding reference number RGPIN-2020-04633]

\bibliography{egbib}
\end{document}